\pdfoutput=1

\documentclass[11pt]{article}

\usepackage[preprint]{acl}

\usepackage{times}
\usepackage{latexsym}

\usepackage[T1]{fontenc}

\usepackage[utf8]{inputenc}

\usepackage{microtype}

\usepackage{inconsolata}

\usepackage{graphicx}

\usepackage{todonotes}

\title{Large-language memorization during the classification of United States Supreme Court Cases}

\author{John E. Ortega \and Dhruv D. Joshi \and Matt P. Borkowski \\
Pace University \\
New York, New York \\
\texttt{Corresponding author: mb26703p@pace.edu}
}

\begin{document}
\maketitle
\begin{abstract}
Large-language models (LLMs) have been shown to respond in a variety of ways for classification tasks outside of question-answering. LLM responses are sometimes called ``hallucinations'' since the output is not what is expected. Memorization strategies in LLMs are being studied in detail, with the goal of understanding how LLMs respond. We perform a deep dive into a classification task based on United States Supreme Court (SCOTUS) decisions. The SCOTUS corpus is an ideal classification task to study for LLM memory accuracy because it presents significant challenges due to extensive sentence length, complex legal terminology, non-standard structure, and domain-specific vocabulary. Experimentation is performed with the latest LLM fine-tuning and retrieval-based approaches, such as parameter-efficient fine-tuning, auto-modeling, and others, on two traditional category-based SCOTUS classification tasks: one with 15 labeled topics and another with 279. We show that prompt-based models with memories, such as DeepSeek, can be more robust than previous BERT-based models on both tasks scoring about 2 points better than previous models not based on prompting.
\end{abstract}

\section{Introduction}

The classification of legal documents can be considered a pivotal task in the streamlining of legal research and decision-making. Wikipedia has already shown that the number of Supreme Court cases in the United States has grown significantly over time. While 7,000-8,000 new cases are filed in the Supreme Court each year, only about 80 of those cases are actually reviewed by the Supreme Court.\footnote{\url{https://en.wikipedia.org/wiki/Number_of_U.S._Supreme_Court_cases_decided_by_year}} With the ever growing demand for legal document processing, one clear task that would assist paralegals and attorneys is the classification of court cases into categorical topics.

The complexity of United States Supreme Court (SCOTUS) decisions presents significant challenges for automated classification systems, as demonstrated by prior work \citep{vatsal2023classification, undavia2018comparative}. Over time, natural language processing (NLP) approaches have evolved—from traditional methods like Latent Dirichlet Allocation (LDA) \citep{blei2003latent} to transformer-based models such as BERT \citep{devlin2018bert}. While BERT is itself a large language model (LLM), previous efforts primarily relied on fine-tuning BERT variants for classification, without a deeper exploration of modern, decoder-based LLMs or prompt-based inference techniques. To our knowledge, this work represents the first formal attempt to classify SCOTUS decisions using a diverse set of contemporary LLM architectures and strategies, including both encoder- and decoder-style models.

In this article, we introduce and compare  four novel LLM components (1) Parameter Efficient Fine-Tuning (PEFT), (2) Prompt-Based Classification (PBC), (3) Automated Model Selection (AutoModel Approach) and (4) Retrieval-Augmented Classification (RAC). Our goal is to better understand memorization and anomalies resulting from the use of these LLM components.

We evaluate the four components on the supreme court database (SCDB), a corpus curated by the University of Washington School of Law that labeled 8.5 thousand court cases into 15 broad categories and 279 detailed sub-categories. Through comprehensive experimentation and the use of accuracy, precision, recall, and F1 score metrics, we aim to identify the most effective approach for classifying SCOTUS decisions across both broad and fine-grained categories.

To better describe our experiments and approach, we first present work that is related to ours in Section \ref{sec:related}. We then provide further insight into the four components studied in Section \ref{sec:methodology}. Next, our work's experimental settings are presented in Section \ref{sec:experimental}. Then, we cover the results of the experiments in Section \ref{sec:results} and provide a final conclusion and future work in Section \ref{sec:conclusion}.

\section{Related Work}
\label{sec:related}

The application of NLP in the legal domain has made significant advancements, particularly with the adoption of transformer-based models like BERT \cite{devlin2018bert}. In the past, BERT models have been fine-tuned for various legal tasks, including document classification, to address challenges posed by the unique characteristics of legal texts, specifically for SCOTUS and SCDB by \citet{vatsal2023classification,undavia2018comparative}.

While the work by \citet{vatsal2023classification,undavia2018comparative} is parallel to ours since it used the same corpora, there was not a deep dive into LLM performance or reasoning as to why model memorization made a difference. Alternatively, others \cite{tewari2024legalpro, chalkidis-etal-2020-legal} introduced \textit{LEGAL-BERT}, a BERT variant pre-trained on a diverse corpus of legal texts that performed ablation studies on various standard NLP tasks like GLUE \cite{wang2018glue} and SQUAD \cite{rajpurkar2016squad}. There work showed how well different approaches maintained robustness across the tasks and, thus, briefly discussed the contributions of BERT.
Nonetheless, the LLMs used in this work were not reviewed.

In another study by \citet{limsopatham-2021-effectively}, legal documents were shown to exceed the token limitations of standard transformer models. Their work explored techniques for adapting BERT to long legal texts, such as hierarchical models and chunking strategies. Their goal was to effectively capture context beyond the typical token limits. In contrast to the work described here, they did not analyze models in the current LLM landscape or decoder-only models.

Some work has been performed that shows how well the integration of retrieval mechanisms with generative models, known as Retrieval-Augmented Generation (RAG), enhance legal text processing. \cite{pipitone2024legalbench} We consider their work more along the lines of ours; yet, there was not a focus on SCOTUS cases as there is here. Similarly,  \citet{wiratunga2024cbr} proposed \textit{CBR-RAG} which combined case-based reasoning with RAG to improve legal question-answering systems.

Other techniques from the literature 
did not do a deep dive into LLM responses 
or did not use the same SCDB tasks based on 15 or 278 categories. This article applies the latest LLM techniques to the SCDB tasks for a classification study to address the challenges of domain-specific language, document length, and the need for precise information retrieval.

\section{Methodology}
\label{sec:methodology}

In this section, we describe the LLMs and approaches (prompt and non-prompt based) that we use in our experiments. There are four models described further in this section: \textbf{BERT} \cite{devlin2019bert}, \textbf{Legal-Bert} \cite{nlpaueb-legalbert}, \textbf{LLaMa 3} \cite{touvron2023llama} and \textbf{DeepSeek} \cite{deepseek2023}. The models presented here incorporate significantly larger parameter counts and diverse architectural innovations compared to earlier approaches \cite{vatsal2023classification, undavia2018comparative} on the same tasks. The overall aim with these four models is to evaluate the impact of memorization, scaling, domain adaptation, and prompt-based inference on the SCDB task. Additionally, in this section we present the three prompt-based approaches and two non prompt-based that we experiment with on the four models. 

\subsection{Models}\label{models-sect}
\label{sec:methodology:models}
We used the following models along with the approaches listed below on each corresponding model:

\textbf{Model 1: BERT.} (Bidirectional Encoder Representations from Transformers): A foundational transformer model pretrained on general English corpora, capturing bidirectional context for various NLP tasks. \cite{devlin2019bert}

\textbf{Model 2: Legal-BERT.} A domain-specific adaptation of BERT, pretrained on extensive legal texts, enhancing its ability to understand legal terminology and nuances. \cite{nlpaueb-legalbert}

\textbf{Model 3: LLaMA 3.} A large language model designed to perform effectively across diverse tasks, including those in specialized domains like law. \cite{touvron2023llama}

\textbf{Model 4: DeepSeek.} A transformer-based large language model, trained on general multilingual corpora, optimized for semantic understanding of prompts, reasoning, and text generation. \cite{deepseek2023}

BERT and Legal-BERT (Models 1 and 2) serve as strong baselines to compare our larger models against. LLaMA and DeepSeek (Models 3 and 4) are up-to-date open-weight LLMs that introduce advanced capabilities through pre-training on massive and diverse corpora and support zero-shot and few-shot generalization.

\subsection{Prompt-Based Approaches}
\label{sec:methodology:prompts}

We experiment with prompt-based approaches with the goal of creating more robust systems. This section enumerates reasons that prompt-based systems may improve results on SCPD tasks. 


\begin{description} 
    \item {\bf Improved Contextual Understanding of Legal Language}: The SCDB  comprises legal documents characterized by complex syntax, formal structure, and domain-specific terminology. Prompt-based large language models (LLMs), such as Model 3, leverage deep bi-directional contextual understanding. We believe that prompting can capture legal nuance more effectively without the need for extensive pre-processing or fine-tuning.   
    \item {\bf Facilitation of Zero-Shot Learning:} Conventional classification models often require large, annotated datasets for supervised learning, which is a limitation in legal domains due to the scarcity and cost of expert-labeled data. Zero-shot learning models, on the other hand, enable classification without task-specific training data. They leverage pre-trained models and specify tasks through natural language prompts. This approach reduces the reliance on labeled data while still delivering high performance across a  range of tasks.
    \item {\bf Domain Transferability and Adaptability:} Prompt-based LLMs, pre-trained on diverse corpora, exhibit strong generalization capabilities across domains. Without requiring task-specific fine-tuning, these models can be easily adapted to sub-fields of law (e.g., constitutional, civil rights, criminal law) through task description embedded in the prompt.
    \item {\bf Semantic Sensitivity to Legal Judgments:} Legal classification tasks often rely on subtle linguistic cues (e.g., distinctions between “affirmed,” “reversed,” or “remanded”). Prompt-based LLMs have demonstrated superior performance in capturing such semantic distinctions through in-context reasoning, as opposed to relying solely on token-level patterns or statistical co-occurrence, which are typical of traditional models.
    \item {\bf Reduced Computational Overhead Compared to Fine-Tuning:} Fine-tuning large-scale models such as Model 2 can be computationally intensive and resource-prohibitive. In contrast, prompt-based methods leverage the inference capabilities of pre-trained models, eliminating the need for additional training.
\end{description}

We implement the following approaches, as motivated in section~\ref{models-sect} for Models 3 and 4. 

\textbf{Approach PB1: Prompt-Based Classification.} This approach leverages the models' ability to perform tasks based on natural language prompts. By crafting specific prompts that guide the model to generate desired outputs, we can effectively utilize the pre-trained knowledge without extensive fine-tuning. Additionally, for large language models such as Model 4, which perform next-token generation as opposed to traditional machine-learning classification, prompting promotes more precise semantic similarity. This method is especially useful in zero-to-few shot learning situations where limited labeled data has a higher demand for in-context learning. This approach is used with \textbf{Model 3}.

\textbf{Approach PB2: Retrieval-Augmented Classification.}  In this approach, we integrate a retrieval mechanism to enhance the models' performance. By retrieving relevant legal documents related to the input text, the model can incorporate additional context, leading to more accurate classifications. This method makes use of external knowledge not included with a pre-trained model's context. We believe that this approach is particularly effective in the legal domain where context and precedent are crucial. This approach is used with \textbf{Model 3}.


\textbf{Approach PB3: Prompt-Based Classification with Log-Smoothed Weighted Loss and Classification Head Adaptation.} This is a variation of Approach PB1, used exclusively for our experiments with Model 4. We attempt to address class and memory imbalance within the SCDB by assigning weights to classes corresponding to their frequency. In practice, certain legal categories tend to appear more frequently than others, we apply logarithmic smoothing of those class weights during fine-tuning. Additionally, to address extreme outlier weights, we introduced a clipping mechanism\footnote{\url{https://pytorch.org/docs/stable/generated/torch.clamp.html}} that will provide an upper bound for class weights. This is more useful for classification than generation by providing logistical probabilities in the response apart from the final classification. This approach is used with \textbf{Model 4}.

In addition the prompt-based approaches mentioned, we define the other approaches below.

\subsection{Non Prompt-Based Approaches}\label{non-prompt-approach-sect}
We also explored several non prompt-based approaches to assess their effectiveness in classifying complex legal texts. These methods are particularly relevant for scenarios where direct prompting may not be ideal due to resource constraints, limited model compatibility, or task-specific performance challenges. Non prompt-based approaches such as Parameter-Efficient Fine-Tuning (PEFT) and Automated Model Selection (AutoModel) allow us to leverage the full capacity of pre-trained models through targeted adaptations and structured evaluation pipelines. By focusing on these techniques, we aim to understand how traditional fine-tuning and optimization pipelines perform in contrast to inference-driven prompting, especially in high-cardinality classification settings like the SCDB dataset. These approaches provide a valuable baseline and comparative benchmark to evaluate the strengths and weaknesses of prompt-based alternatives in both coarse- and fine-grained legal categorization tasks.

\begin{figure*}[t]
\centering
\includegraphics[width=0.87\textwidth]{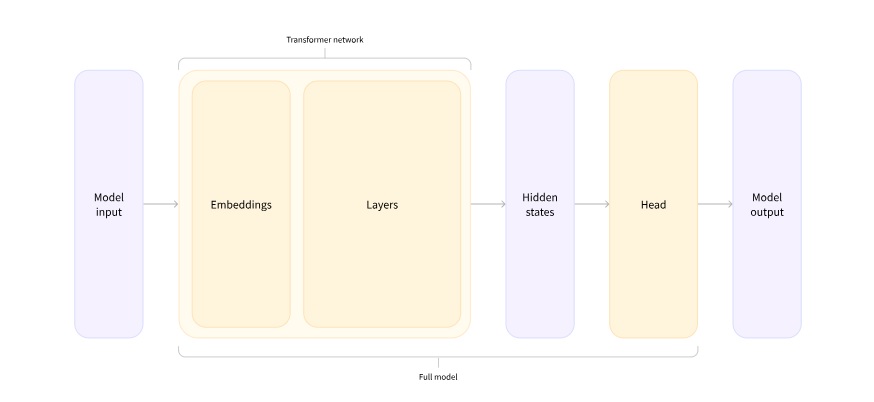} 
\caption{AutoHead model for sequence classification from Hugging Face used for Models 1 and 2.}
\label{sec:methodology:automodel}
\end{figure*}

\textbf{Approach NPB1: Parameter-Efficient Fine-Tuning (PEFT).} We utilize Low-Rank Adaptation (LoRa) \cite{hu2022lora} to fine-tune the models efficiently. LoRA introduces trainable low-rank matrices into each layer of the transformer architecture, allowing adaptation to specific tasks with reduced computational resources. For Models 3 and 4, LoRA helps mitigate the need for full model retraining. This approach is used with \textbf{Model 1}, \textbf{Model 2} and \textbf{Model 4}.


\textbf{Approach NPB2: Automated Model Selection.} We employ an automated framework from HuggingFace\footnote{\url{https://github.com/huggingface/transformers/blob/main/src/transformers/models/auto/modeling_auto.py}} depicted in Figure \ref{sec:methodology:automodel} that selects the most suitable model for classification. It assesses each model's performance on the validation set using metrics such as accuracy, precision, recall, and F1 score. The use of auto-modeling enables an objective comparison and helps identify the model that offers the best trade-off between performance and computational efficiency. \cite{wolf2020transformers} This approach is used with \textbf{Model 1} and \textbf{Model 2}.

\section{Experimental Settings}
\label{sec:experimental}

\subsection{Data}
\label{sec:experimental:data}

The corpus presented here and in previous work \cite{vatsal2023classification, undavia2018comparative} is based on the SCDB a publicly available dataset curated by Washington University School of Law. \cite{spaeth2013supreme} The corpus consists of 8,419 Supreme Court decisions, each labeled with a primary issue area. SCDB provides a two-level ontology for classification: (i) 15 broad categories representing high-level legal domains (e.g., Civil Rights, Criminal Procedure, Economic Activity) and (ii) 279 fine-grained categories that further specify legal subtopics within each broad category. The class distribution can be considered imbalanced across categories. Documents length can be considered highly variable, with a median length of 5,552 tokens and a mean length of 6,960 tokens, posing challenges for models with input length restrictions (e.g., 512-token limit in BERT).

\subsection{Chunking}
\label{sec:experimental:Chunking}

Our chunking strategies for the approaches in sections~\ref{sec:methodology:prompts} and \ref{non-prompt-approach-sect}
are as follows:
\begin{description}
    \item \textbf{NPB2. Models 1 and 2}: Tokenization is performed with the HuggingFace AutoTokenizer where max length is set to 512 and truncation is set. We apply batch-wise processing and texts that exceed 512 tokens are truncated.
    \item \textbf{NPB2. Models 1 and 2}: The use of transformer-based classification and chunking with a max length of 512; batches are of size 4.
    \item \textbf{PB1. Model 3}: Full court case texts are used that do no exceed the context limit ($\approx$2048-4096 tokens).
    \item \textbf{PB2. Model 3}: Chunking is used based on a dictionary. Text is segmented according to the structure of the document.
    \item \textbf{PB3. Model 4}: Chunks of the document are truncated according to the model's context window (up to 128,000 tokens) and resources on hand. In our experiments, we chose a limit of 5,000 tokens.
\end{description}

\section{Results and Discussion}
\label{sec:results}

Our results for both categories from the SCDB corpus are shown in Table \ref{results15table} for 15 broad categories and in Table \ref{results279table} for 279 detailed categories.


It is clear that Model 4 is the most robust overall for both tasks, despite it scoring slightly lower (1 misclassification) on 279 categories. Additionally, Models 1 and 2 have outperformed previous work \cite{vatsal2023classification, undavia2018comparative}. Although, the used of the NPB2 technique in Model 2 has outperformed Model 4 in the categorization of 279 labels. This dichotomy between BERT-based models and largely prompt-based models is interested to note since these experiments are performed on complex legal structure. Model 3 did not perform well on the 15 broad categories compared to the other models. The memory consumption overall caused several issue due to the the Ollama framework which was abandoned for the 279 categorical task despite several attempt to restart and address the issues. We plan on addressing this for future work but are convinced that its low performance on 15 categories would only be replicated for 279.

\begin{table*}
\centering

\begin{tabular}{ ccccc }
\hline
\bfseries{Model} & \bfseries {Technique} & \bfseries{Accuracy}  & \bfseries{Precision}  & \bfseries{F1}  \\
  
\hline
  & NPB1 & 0.718  & 0.669   & 0.679\\
Model 1 & NPB2 & 0.785  & 0.779   & 0.780\\

\hline
 & NPB1 & 0.748 & 0.705   & 0.715\\
Model 2 & NPB2 & 0.793  & 0.792   & 0.791\\
\hline
 & PB1 & 0.554 & 0.633   & 0.577\\
Model 3 & PB2 & 0.332  & 0.461   & 0.346\\
\hline
  Model 4 & PB3 &  \textbf{0.824}  & \textbf{0.827}  & \textbf{0.820} \\
 \hline
\end{tabular}
\caption{Results For 15 Categories}
\label{results15table}
\end{table*}

\begin{table*}
\centering
\begin{tabular}{ ccccc }
\hline
\bfseries{Model} & \bfseries {Technique} & \bfseries{Accuracy}  & \bfseries{Precision}  & \bfseries{F1}  \\
\hline
  
  & NPB1 & 0.171  & 0.042   & 0.063\\ 

Model 1  & NPB2 & 0.581  & 0.542   & 0.541\\
\hline

  & NPB1  & 0.365 & 0.201 & 0.246\\

Model 2  & NPB2 & \textbf{0.624}  & 0.617   & \textbf{0.602}\\
\hline
Model 4 & PB3  &  0.620  & \textbf{0.624}  & 0.598 \\
\hline

\end{tabular}
\caption{Results For 279 Categories}
\label{results279table}

\end{table*}

Contrastingly, the approaches for Model 3 were significantly different. PB2 (a retrieval-augmented generation (RAG) approach) achieves a classification accuracy of 0.55 compared to just 0.33 from PB1 (a prompt-based approach). This performance discrepancy can be attributed primarily to the differences in architectural focus and task alignment between the two methods. PB1's prompting strategy was tailored for direct classification where the model receives structured input and is explicitly guided to produce a classification label. In contrast, PB2, while powerful for open-domain QA and generative reasoning, introduces multiple points of potential failure in a classification setting. Its reliance on a retrieval step followed by generation introduces noise, especially when the retrieved documents are not perfectly aligned with the classification target. These results suggest that while RAG is well-suited for tasks requiring long-context understanding and answer synthesis, prompt-based LLM classification is more effective for structured prediction tasks like legal case categorization.

Table \ref{results15table} provides the \textit{best} results for Models 1 and 2 (both BERT-based) on 15 categories after trying several sequence modeling techniques. It is noteworthy that the best results for both models were gotten using the Stride-64 n-gram compression technique experimented with in previous work \cite{vatsal2023classification,undavia2018comparative}. Further results on both models are in order of best-to-worst performance as such: Concat-512, Summarization-512, Best-512, Ensemble and  LSMs (Longer Sequence Models). We believe that there could be multiple reasons that the LSM technique underperformed in our experiments. First, LSMs allow multi-head attentions to adhere to a restricted \textit{memorization} window contrary to BERT-based models where these multi-head attentions are free to concentrate on any of the tokens. Second, with more number of tokens, more variance is created which can lead to poor performance due to the wider search space. In our studies, we find that the first 512 tokens contribute the most and this \textit{may} be attributed to the fact that the documents are of unequal length and there are many documents which are less than or equal to the length of 512 tokens. Additionally, Best-512 probably performs worse than Concat-512 and Best-512 because, even though the first 512 tokens are the most important, context beyond these 512 tokens still contributes and should not be ignored. Summarization-512 reduction over an entire document is limited due to its attempt to narrow down larger documents to 512 tokens. As expected, the ensemble method outperforms Best-512 and Summarization-512 because it tries to exploit joint learning across multiple 512 token chunks through its maximum voting mechanism rule. Concat-512 on the other hand captures better context across multiple 512 token chunks as it learns this knowledge during back propagation of the model. Finally, Stride-64 outperforms Concat-512 because when we take disjoint chunks, the continuity of context goes missing whereas if there is an overlapping portion of text between two consecutive chunks, it gives better contextual understanding.

Our evaluation demonstrates that LegalBERT consistently outperforms the standard BERT model across both coarse-grained (15 categories) and fine-grained (279 categories) legal classification tasks. In the 15-class task, while BERT showed reasonable performance, LegalBERT delivered noticeably better results on both evaluation datasets, reflecting its ability to better capture the structure and semantics of legal texts. The improvement is even more pronounced in the 279-class classification setting, where the complexity and specificity of labels make domain understanding critical. BERT’s performance dropped significantly in this scenario, particularly on more challenging datasets. LegalBERT, however, showed greater robustness and maintained relatively strong performance, confirming the value of domain-specific pretraining in legal NLP. Overall, these results highlight that LegalBERT is more effective and reliable for legal case classification, especially when dealing with nuanced or large-scale categorization tasks.

We observed a significant drop in classification performance when increasing the number of output labels from 15 to 279, despite using identical model architectures and training settings across both BERT and LoRA-adapted BERT (via PEFT). This degradation can be attributed to multiple compounding factors inherent to high-cardinality classification tasks. First, with a large number of classes, the final softmax layer distributes probability mass across more labels, resulting in reduced confidence and increased prediction ambiguity. Second, the effective sample size per class drops sharply, leading to high variance and poor generalization, especially for underrepresented labels. Crucially, LoRA adapters modify internal attention layers but do not adapt the classification head, which becomes a significant bottleneck in high-dimensional label spaces. The randomly initialized classification head (e.g., 768×279) lacks sufficient supervision for effective optimization without full fine-tuning. Furthermore, traditional cross-entropy loss becomes less effective in such settings, particularly when class distributions are imbalanced. These observations were consistent in both standard BERT fine-tuning and LoRA-based setups, suggesting a fundamental challenge related to label-space sparsity rather than parameter adaptation strategy. Addressing this requires techniques such as classifier head augmentation, label smoothing, or hierarchical label modeling. Interestingly, models instantiated using Hugging Face’s AutoModelForSequenceClassification tend to yield better performance in such settings. This is because the class encapsulates task-specific heads with appropriate initialization schemes, dropout strategies, and architectural configurations optimized for classification. It ensures consistency with pretraining objectives and provides robust handling of class imbalance and label index alignment internally. Moreover, the AutoModel abstraction leverages model-specific nuances (e.g., classification token treatment in BERT vs. LegalBERT), ensuring better downstream performance across diverse architectures with minimal manual adjustments.

The limitations of BERT-based models due to context-window size points to a conclusion that prompt-based LLMs could outperform them when measuring accuracy, precision and recall. However, the possibility of a smaller model, such as Model 2, with a lower memory footprint and availability of on-premise inference is also promising. It is clear to us that larger models that allow for more in-context memory and, thus, larger context windows with refined outputs provide better results. Model 4, based on DeepSeek R1, is able to take advantage of context as much as 10 times larger than Models 1 and 2 in our experiments. Model 3, as previously explained, has limitations and, due to its overly large context-window size and other memorization attributes, was rendered unusable in our experiments given the technology on hand, two Nvidia Quadro P4000 with 8 GB of memory each. This, thus, implies that in our experiments Model 4 is the best performing model due to its robustness for both tasks.

\section{Conclusion \& Future Work}
\label{sec:conclusion}

Recent advances in legal NLP have highlighted the challenges of classifying long-form documents such as U.S. Supreme Court decisions. One of the most notable works in this area is by Vatsal et al. (2023) \cite{vatsal2023classification}, who proposed using BERT-based techniques for classifying such cases. Due to the inherent 512-token limitation of BERT, their approach employed chunking and summarization to preprocess the legal texts. By applying ensemble voting over multiple chunked or summarized inputs, they achieved commendable results: 0.8 accuracy for a 15-class classification task and 0.6 accuracy on a more fine-grained 279-class categorization task.
Our methodology diverged from Vatsal et al.’s \cite{vatsal2023classification} in several key ways. By using LegalBERT, a model pretrained on legal corpora, we were able to better capture semantic and syntactic nuances of legal language. Additionally, instead of relying solely on chunk-based voting or summarization, our approach preserved more of the document’s context—either by optimizing how chunks were processed or by better leveraging attention mechanisms within the model. Our training pipeline also incorporated advanced scheduling, regularization, and evaluation, leading to more robust generalization for comparison purposes.

These findings demonstrate that domain-adapted models combined with parameter-efficient fine-tuning can provide state-of-the-art performance in legal case classification. Our approach not only improves accuracy over traditional chunk-based strategies but also offers greater scalability and efficiency—making it a compelling solution for real-world legal AI systems.

As a part of future work, we can leverage the knowledge embedded in references of a given SCDB document. A reference in an SCDB document basically refers to some other SCDB document that has been cited to legally justify the decision taken on the former SCDB document. The raw text of these references may not be very helpful in improving the classification tasks. We can use a graph structure to denote the relations between a given SCDB document with other documents cited in it. The final classification result of a given SCDB document can be calculated based on some form of aggregation incorporating classification results of its references weighted by the graph structure representing quantified relations with the SCDB document at hand.

Another area of future work can be applying greedy approaches to the techniques discussed in this work. For example, we can have a greedy summarization technique where in the final 512 token summary, we can include more number of tokens from the best performing 512 token chunk as inferred from Best-512 technique. Similarly, we can have greedy ensemble technique where during the voting phase, we can give more weight to the best performing 512 token chunk as the results show from Best-512 technique.

\bibliography{acl}
\appendix
\section{Prompts}
\label{se:appendix:prompts}
{\footnotesize

\subsection{Model 4 (Deep Seek)}

The following is a U.S. Supreme Court opinion. Your task is to classify it into one of the 263 known legal categories based on its subject matter.

{Important Notes:}
\begin{itemize}
    \item The labels are numeric (e.g., 0-262) but {do not} indicate a ranking or relationship between categories. Instead, the labels represent various legal topics, examples of which include: `employment discrimination based on race, age, religion, or national origin', `affirmative action', `civil rights protests', and `First Amendment - free speech'.
    \item Carefully read the opinion and determine the {most appropriate} category.
    \item If the opinion discusses {multiple legal issues}, assign the {primary} category.
    \item {Note:} These opinions have been {truncated} for brevity but retain essential context.
\end{itemize}

{Example Classifications:}

Opinion (truncated): "Petitioner, an Orthodox Jew and ordained rabbi, was ordered not to wear a yarmulke while on duty and in uniform as a commissioned officer in the Air Force at March Air Force Base, pursuant to an Air Force regulation that provides that authorized headgear may be worn out of doors but that indoors "[h]eadgear [may] not be worn ... except by armed security police in the performance of their duties." Petitioner then brought an action in Federal District Court, claiming that the application of the regulation to prevent him from wearing his yarmulke infringed upon his First Amendment freedom to exercise his religious beliefs. The District Court permanently enjoined the Air Force from enforcing the regulation against petitioner. The Court of Appea..."
Category: (115)

Opinion (truncated): "Respondent was convicted of second-degree murder for killing one Sewell with a knife during a fight. Evidence at the trial disclosed, inter alia, that Sewell, just before the killing, had been carrying two knives, including the one with which respondent stabbed him, that he had been repeatedly stabbed, but that respondent herself was uninjured. Subsequently, respondent's counsel moved for a new trial, asserting that he had discovered that Sewell had a prior criminal record (including guilty pleas to charges of assault and carrying a deadly weapon, apparently a knife) that would have tended to support the argument that respondent acted in self-defense, and that the prosecutor had failed to disclose this information to the defense. The Distri..."
Category: (120)

Opinion (truncated): "To assist airline employees dislocated as a result of the deregulation of commercial air carriers pursuant to the Airline Deregulation Act of 1978 (Act), Congress enacted an Employee Protection Program (EPP) as 43 of the Act. The EPP imposes on covered airlines the "duty to hire" dislocated protected employees, who have a "first right of hire" in their occupational specialities with any covered airline that is hiring additional employees. Section 43 authorizes the Secretary of Labor to issue regulations for the administration of the EPP, but 43(f)(3) contains a legislative-veto provision stating that any final regulation shall become effective after 60 legislative days following its submission to Congress, unless during that period either H..."
Category: (259)

Opinion (truncated): "In petitioner's employment compensation action, which respondent removed from a Colorado state court to the Federal District Court on the basis of diversity of citizenship, judgment was entered on the jury's verdict for petitioner in an amount considerably less than he had sought. Petitioner timely filed new-trial motions and a motion for attorney's fees under Colorado law. On May 14, 1984, the court denied the new trial motions but found that petitioner was entitled to attorney's fees, and, on August 1, 1984, entered a final order determining the amount of the fees. On August 29, petitioner filed notice of appeal to the Court of Appeals, covering all of the District Court's post-trial orders. Although affirming the attorney's fees award, t..."
Category: (226)

Opinion (truncated): "[
Footnote *
] Together with No. 89-700, Astroline Communications Company Limited Partnership v. Shurberg Broadcasting of Hartford, Inc., et al., also on certiorari to the same court.
 These cases consider the constitutionality of two minority preference policies adopted by the Federal Communications Commission (FCC). First, the FCC awards an enhancement for minority ownership and participation in management, which is weighed together with all other relevant factors in comparing mutually exclusive applications for licenses for new radio or television broadcast stations. Second, the FCC's so-called "distress sale" policy allows a radio or television broadcaster whose qualifications to hold a license have come into question to transfer that l..."
Category: (66)

{Now classify the following opinion:}

--- Opinion Start ---

--- Opinion End ---

Provide the correct category number.
\subsection{Deep Seek}
\label{se:appendix:prompts:se:deep}
The following is a U.S. Supreme Court opinion. Your task is to classify it into one of the 13 known legal categories based on its subject matter.

Important Notes:

- The labels are numeric (e.g., 0-12) but do not indicate a ranking or relationship between categories. Instead, the labels represent various legal topics:

0: Criminal Procedure,

1: Civil Rights,

2: First Amendment,

3: Due Process,

4: Privacy,

5: Attorneys,

6: Unions,

7: Economic Activity,

8: Judicial Power,

9: Federalism,

10: Interstate Relations,

11: Federal Taxation,

12: Miscellaneous

- Carefully read the opinion and determine the most appropriate category.

- If the opinion discusses multiple legal issues, assign the primary category.

-  Note: These opinions have been truncated for brevity but retain essential context.

 Example Classifications:
 
Opinion (truncated): "In an action brought to restrain the enforcement of a state statute on federal constitutional grounds, the federal court should retain jurisdiction until a definitive determination of local law question is obtained from the local courts; and the judgment of the District Court in this case is vacated and the cause is remanded to it with directions to retain jurisdiction until efforts to obtain an appropriate adjudication in the state courts have been exhausted. Pp. 364-367.
 146 F. Supp. 214, judgment vacated and cause remanded with directions.
Milton I. Shadur argued the cause for appellants. With him on the brief were Arthur J. Goldberg, David E. Feller and Herbert S. Thatcher.
Gordon Madison, Assistant Attorney General of Alabama, argued ..."
Category: (9)

Opinion (truncated): "Rehearing Denied March 3, 1947
 See . Messrs. Ford W. Thompson and J. L. London, both of St. Louis, Mo., for appellant.
 Mr. Ferre C. Watkins, of Chicago, Ill., for appellee.
 Mr. Justice DOUGLAS delivered the opinion of the Court.
 This case presents a substantial question under the Full Faith and Credit Clause (Art. IV, 1) of the Constitution.
 Chicago Lloyds, an unincorporated association, was authorized by Illinois to transact an insurance business in Illinois and other States. It qualified to do business in Missouri. In 1934 petitioner sued Chicago Lloyds in a Missouri court for malicious prosecution and false arrest. In 1938, before judgment was obtained in Missouri, respondent's predecessor was appointed by an Illinois court as statu..."
Category: (11)

Opinion (truncated): "620 F.2d 17, affirmed by an equally divided Court.
Mark I. Levy argued the cause for petitioners. With him on the briefs were Solicitor General McCree, Acting Assistant Attorney General Martin, Deputy Solicitor General Geller, Michael Kimmel, Robert Kaplan, and Milton J. Socolar.
Gilbert H. Weil argued the cause for respondent. With him on the brief was Robert L. Sherman.
*
 [
Footnote *
] Philip Lacovara, Ronald A. Stern, Stanley L. Temko, Charles Lister, and Clare Dalton filed a brief for Merck \& Co., Inc., et al. as amici curiae urging affirmance.
PER CURIAM.
The judgment is affirmed by an equally divided Court.
JUSTICE STEWART took no part in the consideration or decision of this case...."
Category: (0)

Opinion (truncated): "To avert a nation-wide strike of steel workers in April 1952, which he believed would jeopardize national defense, the President issued an Executive Order directing the Secretary of Commerce to seize and operate most of the steel mills. The Order was not based upon any specific statutory authority but was based generally upon all powers vested in the President by the Constitution and laws of the United States and as President of the United States and Commander in Chief of the Armed Forces. The Secretary issued an order seizing the steel mills and directing their presidents to operate them as operating managers for the United States in accordance with his regulations and directions. The president promptly reported these events to Congress; b..."
Category: (13)

Opinion (truncated): "A Georgia statute requires candidates for designated state offices to certify that they have taken a urinalysis drug test within 30 days prior to qualifying for nomination or election and that the test result was negative. Petitioners, Libertarian Party nominees for state offices subject to the statute's requirements, filed this action in the District Court about one month before the deadline for submission of the certificates. Naming as defendants the Governor and two officials involved in the statute's administration, petitioners asserted, inter alia, that the drug tests violated their rights under the First, Fourth, and Fourteenth Amendments to the United States Constitution. The District Court denied petitioners' motion for a preliminar..."
Category: (5)

Now classify the following opinion:

--- Opinion Start ---

--- Opinion End ---

Provide the correct category number.
}

\subsection{Model 3 (LLaMa 3)}
\label{se:appendix:prompts:se:ollama}
\textbf{Prompts for Approach PB1 for Model3}

\textbf{System prompt:}
You are an expert in Supreme Court cases and legal classifications. Your task is to analyze the provided text of a Supreme Court case and classify it into one of the following predefined categories:

Instructions:

1. Carefully analyze the text and identify the primary category it belongs to.

2. Provide only the category name as the response and nothing else.

3. If uncertain, select the most relevant category based on the case's primary legal focus.

4. Response should not exceed more than 2 words.

\textbf{User prompt:}
(case text using f string)

Please reply with the category only and nothing else. Ensure its category is from the given list or the most relevant one. Here are the categories:

- None: Cases that do not fit into any specific category.

- Criminal Procedure: Cases involving the process of investigating and prosecuting crimes.

- Civil Rights: Cases about the rights of individuals to receive equal treatment.

- First Amendment: Cases addressing issues related to freedom of speech, religion, or press.

- Due Process: Cases focusing on legal safeguards ensuring fair treatment.

- Privacy: Cases dealing with the right to privacy.

- Attorneys: Cases about the legal profession and lawyer regulations.

- Unions: Cases concerning labor unions and collective bargaining.

- Economic Activity: Cases involving business, trade, or commerce.

- Judicial Power: Cases addressing the powers of courts and the judiciary.

- Federalism: Cases about the division of power between state and federal governments.

- Interstate Relations: Cases about interactions between states.

- Federal Taxation: Cases involving federal tax laws.

- Miscellaneous: Cases that do not clearly fit into any of the above categories.

- Private Action: Cases involving disputes between private individuals or entities.
\newline

\textbf{Prompts for Approach PB2 for Model3}

\textbf{Initial prompt for reference data:}
Give me a list of words and phrases (atleast 15) which would be present in a case text to classify it into following categories. Do not write any explaination or warnings.

- Criminal Procedure: Cases involving the process of investigating and prosecuting crimes.

- Civil Rights: Cases about the rights of individuals to receive equal treatment.

- First Amendment: Cases addressing issues related to freedom of speech, religion, or press.

- Due Process: Cases focusing on legal safeguards ensuring fair treatment.

- Privacy: Cases dealing with the right to privacy.

- Attorneys: Cases about the legal profession and lawyer regulations.

- Unions: Cases concerning labor unions and collective bargaining.

- Economic Activity: Cases involving business, trade, or commerce.

- Judicial Power: Cases addressing the powers of courts and the judiciary.

- Federalism: Cases about the division of power between state and federal governments.

- Interstate Relations: Cases about interactions between states.

- Federal Taxation: Cases involving federal tax laws.

- Miscellaneous: Cases that do not clearly fit into any of the above categories.

- Private Action: Cases involving disputes between private individuals or entities.

Format the output in the following way:

'Category': words, phrase

\textbf{Prompt for classification:}
You are an AI assistant trained in Supreme Court case classification. Given the following retrieved categories,
classify the new case into the most appropriate issue area.

Retrieved Relevant Categories:

(retrieved text from RAG)

Please reply with the category only and nothing else.

\end{document}